\documentclass[letterpaper]{article} 
\usepackage{aaai23}  
\usepackage{times}  
\usepackage{helvet}  
\usepackage{courier}  
\usepackage[hyphens]{url}  
\usepackage{graphicx} 
\usepackage{natbib}  
\usepackage{caption} 
\usepackage{algorithm}
\usepackage{algorithmic}
\usepackage{enumitem}

\usepackage{newfloat}
\usepackage{listings}
\DeclareCaptionStyle{ruled}{labelfont=normalfont,labelsep=colon,strut=off} 
\floatstyle{ruled}
\newfloat{listing}{tb}{lst}{}
\floatname{listing}{Listing}
\title{Contrastive View Design Strategies to Enhance Robustness to Domain Shifts in Downstream Object Detection}
\author{
    Kyle Buettner\textsuperscript{\rm 1},
    Adriana Kovashka\textsuperscript{\rm 2}
}
\affiliations{
    \textsuperscript{1}Intelligent Systems Program, 
    \textsuperscript{2}Department of Computer Science \\
    University of Pittsburgh, PA, USA\\

    buettnerk@pitt.edu, kovashka@cs.pitt.edu
}

\newcommand{\eg}{\textit{e}.\textit{g}. }

\usepackage{bibentry}

\usepackage{xcolor,colortbl}

\definecolor{LGray}{gray}{0.95}
\definecolor{Gray}{gray}{0.87}
\definecolor{LightCyan}{rgb}{0.88,1,1}

\newcolumntype{a}{>{\columncolor{Gray}}c}
\newcolumntype{g}{>{\columncolor{LGray}}c}

\begin{document}

\maketitle

\begin{abstract}
      Contrastive learning has emerged as a competitive pretraining method for object detection. Despite this progress, there has been minimal investigation into the robustness of contrastively pretrained detectors when faced with domain shifts. To address this gap, we conduct an empirical study of contrastive learning and out-of-domain object detection, studying how contrastive view design affects robustness. In particular, we perform a case study of the detection-focused pretext task Instance Localization (InsLoc) and propose strategies to augment views and enhance robustness in appearance-shifted and context-shifted scenarios. Amongst these strategies, we propose changes to cropping such as altering the percentage used, adding IoU constraints, and integrating saliency-based object priors. We also explore the addition of shortcut-reducing augmentations such as Poisson blending, texture flattening, and elastic deformation. We benchmark these strategies on abstract, weather, and context domain shifts and illustrate robust ways to combine them, in both pretraining on single-object and multi-object image datasets. Overall, our results and insights show how to ensure robustness through the choice of views in contrastive learning. 
\end{abstract}

 \section{Introduction}
    
            Self-supervised learning has been rising in popularity in computer vision, with many top methods using contrastive learning \cite{hadsell2006dimensionality}, a form of learning that optimizes feature representations for positive samples to be close together and for negative samples to be far apart. Self-supervised contrastive models such as SimCLR \cite{chen2020simple} and MoCo \cite{he2020momentum} have been shown to approach or surpass the performance of supervised models when representations are transferred to downstream image classification, object detection, and semantic segmentation tasks \cite{ericsson2021well}. This success is in part due to strategic data augmentation pipelines that these models use to create effective positive and negative \textit{views} (samples) for learning. 
        
\begin{figure}[t]
    \centering
    \includegraphics[scale=0.26]{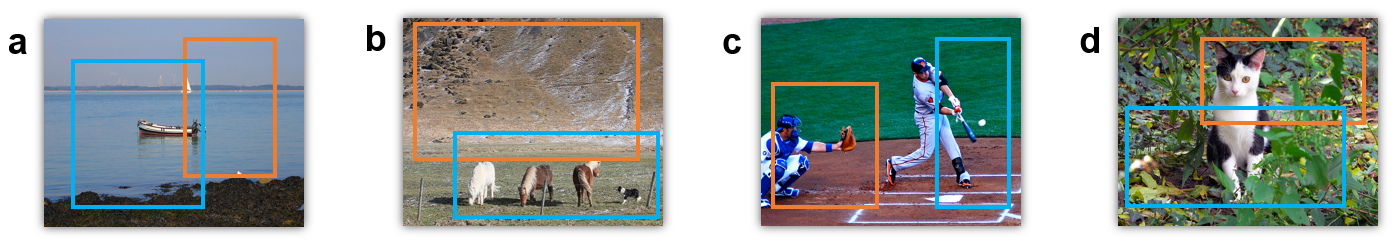}
    \caption{\textbf{What properties of contrastive views enable robust downstream object detection?} Existing contrastive view design methods may cause detectors to lack robustness in domain-shifted settings (\eg appearance, context). For instance, state-of-the-art pipelines use random cropping, which may cause contextual bias as objects can be aligned to common backgrounds (\eg boat in \textbf{a}, horses in \textbf{b}) or co-occurring objects (\eg player, glove, ball, and bat in \textbf{c}). Due to a lack of spatial consistency between views, its use may also discourage the learning of object shape  (\eg cat in  \textbf{d}).}
    \label{intro_fig}
\end{figure}

            Despite this progress, the out-of-distribution robustness of contrastive representations has been minimally studied, especially with regards to object detection. We hypothesize that existing data augmentation pipelines in contrastive learning may result in representations that \textit{lack} robustness in such domain-shifted detection scenarios. For example, as shown in Fig. \ref{intro_fig}, state-of-the-art pipelines (\eg SimCLR, MoCo) create positive views with aggressive random cropping of a single image. Such use of this augmentation can lead to features for object regions being made similar to those for the background or co-occurring objects, potentially causing contextual bias and hurting in out-of-context scenarios. Random cropping may also result in texture-biased rather than shape-biased features, since the shapes of object parts may not be consistent across crops. A lack of shape in representations can lead to degraded performance when texture is shifted (\eg appearance changes due to weather).
            
            In this work, we explore strategies to improve contrastive view design for enhanced robustness in domain-shifted detection. In particular, we conduct an empirical study of the state-of-the-art detection pretext task InsLoc \cite{yang2021instance} and strategically alter the views created from InsLoc's data augmentation pipeline by adjusting cropping, adding shortcut-reducing appearance augmentations, and integrating saliency-based object priors. These strategies are evaluated in-domain and out-of-domain in appearance-shifted and context-shifted detection scenarios. Experiments are also conducted following pretraining on both single-object and multi-object image datasets. From these experiments, we present these insights into contrastive view design:

            \begin{itemize}[itemsep=1mm, parsep=0pt]
                  \item Increasing the minimum \% of an image used for crops or adding an IoU constraint between views causes the model to learn from spatially consistent views with larger object parts. These strategies encourage the learning of shape and improve robustness to appearance shifts.
                  \item Shortcut-reducing augmentations enhance the effectiveness of non-aggressive cropping, exemplified by improvements over InsLoc both \textit{in-domain} (up to +2.73 AP) and \textit{out-of-domain} (up to +3.07 AP).
                  \item The use of saliency priors in views is effective for out-of-context robustness. Their use is best in a mechanism that removes background and tightens crops to object regions.
                  \item Applying shortcut-reducing augmentation to the non-salient regions in views, in combination with crop tightening and shape strategies, is effective for enhancing robustness to both appearance and context shifts.
            \end{itemize}
        
    \section{Background and Related Work}
    
        \subsubsection{Self-supervised and contrastive learning} Many top self-supervised methods in vision use contrastive learning and the instance discrimination pretext task \cite{wu2018unsupervised}, where each image is its own class, and the goal is to discern that two positive samples (or \textit{views}) are from the same image when considered versus a set of negatives. Typically, positives are generated through aggressive data augmentation of a single image, and they are compared to a large number of negatives from other images. While the representations for negatives have been stored in large memory banks \cite{wu2018unsupervised}, recent methods optimize the learning pipeline to rather use large batch sizes \cite{chen2020simple} or a dynamic dictionary \cite{he2020momentum}. Alternatively, the online clustering approach of \cite{caron2020unsupervised} avoids the need for pairwise comparisons entirely, and the iterative approach of BYOL \cite{grill2020bootstrap} avoids using negatives. In general, contrastive methods are evaluated by transferring representations to downstream tasks such as object detection, and \textit{domain shifts are not usually considered with detection}. General detectors have been shown to lack robustness to shifts, retaining only 30-60\% of performance when tested on natural corruptions \cite{michaelis2019benchmarking}, and contrastive detectors may similarly lack robustness. In this work, we fill in the need to more broadly characterize and improve the generalizability of contrastive representations through our study of view design strategies in domain-shifted detection. 
        \subsubsection{Data augmentations and views in contrastive learning} Recent works have investigated how to construct positive and negative views for contrastive learning. In particular, topics explored include how to learn views \cite{tian2020makes}, how to mine hard negatives \cite{kalantidis2020hard, robinson2020contrastive}, how to handle false negatives \cite{chuang2020debiased}, and how to modify sample features to avoid shortcuts \cite{robinson2021can}. Positives have also been studied with regards to intra-image augmentations and instance discrimination. For instance, SimCLR \cite {chen2020simple} finds that creating positives with random cropping, color distortion, and Gaussian blur is effective for ImageNet classification. In this work, we also empirically explore augmentations for positives, but consider those specifically targeting domain robustness in detection, \textit{which include some previously unexplored in contrastive learning: Poisson blending, texture flattening, and elastic deformation}. We also show that modifying cropping to be less aggressive or use IoU constraints can improve out-of-domain robustness.
        
        
        

        
        \subsubsection{Robustness in contrastive learning} Neural network representations have been shown to not generalize well to various domain shifts (\eg pose \cite{alcorn2019strike}, corruptions \cite{hendrycks2019benchmarking}) and to suffer from biases (\eg texture \cite{geirhos2018imagenet}, context \cite{singh2020don}, background \cite{xiao2020noise}). Contrastive representations face similar issues, for instance versus viewpoint shifts \cite{purushwalkam2020demystifying} and texture-shape conflicts \cite{geirhos2020surprising}. \textit{Most works that strive to explicitly improve contrastive robustness either focus on image recognition \cite{ge2021robust, khosla2020supervised}, proxy recognition tasks like the Background Challenge \cite{xiao2020noise}, or evaluate only when transferring representations to object detection, but not on domain shifts in detection.} We alternatively consider contrastive robustness with respect to object detection and relevant domain shifts. One shift we consider is \textit{in context}, as research has identified contextual bias as an issue in contrastive pretraining on multi-object image datasets. For example, \cite{selvaraju2021casting} addresses contextual bias in COCO pretraining by constraining crops to overlap with saliency maps and using a Grad-CAM attention loss, leading to improvements over MoCo on COCO and VOC detection. Similarly, \cite{mo2021object} proposes two augmentations, object-aware cropping (OA-Crop) and background mixup (BG-Mixup), to reduce contextual bias in MoCo-v2 and BYOL. Notably, these cropping strategies have minimally been tested with detection (just in-domain), so it is unclear how such strategies perform in out-of-domain detection. We evaluate these strategies out-of-domain and show that they do not always result in  improvements. We thus propose a hybrid strategy of the methods and show that it substantially improves out-of-context robustness.


        \subsubsection{Pretext tasks for object detection} Our research also fits with recent works that tailor pretraining to downstream tasks besides image classification, such as object detection. In particular, we explore InsLoc, a pretext task in which detection-focused representations are built in contrastive learning through integration of bounding box information \cite{yang2021instance}. Other notable approaches exist which leverage selective search proposals \cite{wei2021aligning}, global and local views \cite{Xie_2021_ICCV}, spatially consistent representation learning \cite{roh2021spatially}, weak supervision \cite{Zhong_2021_CVPR}, pixel-level pretext tasks \cite{wang2021dense,xie2021propagate}, and transformers \cite{dai2021up}. \textit{Our work is orthogonal to such works as we provide contrastive view design insights that can guide future detection-focused pretext tasks to have greater out-of-domain robustness.}

    \section{Experimental Approach}
    
        In this study, our goal is to analyze how strategic changes to contrastive views impact downstream object detection robustness.
        In particular, we consider two families of domain shifts that cause drops in object detection performance: appearance and context. \emph{Appearance shift} in our study is defined as change in the visual characteristics of objects (such as color brightness and texture). \emph{Context shift} in our study is defined as when an object appears with different objects or in different backgrounds during train and test time. 
       
        Formally, we view contrastive learning from a knowledge transfer perspective. There is a source (pretext) task \textit{$s$} and a downstream object detection task \textit{$d$}. Unsupervised, contrastive pretraining is performed in \textit{$s$} on dataset $\mathbf s$, then representations are transferred to \textit{$d$} for supervised finetuning on dataset $\mathbf d$. We explore strategies to adapt \textit{$s$} to \textit{$s'$}, which represents a task with views augmented to be robust to families of domain shifts (\eg appearance, context). To evaluate the effectiveness of \textit{$s'$}, representations from $s'$ (after finetuning on $\mathbf{d}$) are evaluated on $\mathbf{d}$ and various $\mathbf{t_i}$, which represent target datasets domain-shifted from $\mathbf d$.

          \subsubsection{Base pretext task}
            
            We select \textit{s} to be Instance Localization (InsLoc) \cite{yang2021instance}, a state-of-the-art detection-focused, contrastive pretext task. In particular, InsLoc is designed as an improvement over MoCo-v2 \cite{chen2020improved} and uses the Faster R-CNN detector \cite{ren2015faster}. Positive views are created by pasting random crops from an image at various aspect ratios and scales onto random locations of two random background images. Instance discrimination is then performed between foreground features obtained with RoI-Align \cite{He_2017_ICCV}  from the two composited images (positive \textit{query} and \textit{key} views) and negatives maintained in a dictionary. This task is optimized with the InfoNCE loss \cite{van2018representation}.

            Outlining the specifics of InsLoc's augmentation pipeline further, the crops used for views are uniformly sampled to be between 20-100\% of an image. Crops are then resized to random aspect ratios between 0.5 and 2 and width and height scales between 128 and 256 pixels. Composited images have size 256$\times$256 pixels. Other augmentations used include random applications of Gaussian blurring, horizontal flipping, color jittering, and grayscale conversion. Notably, InsLoc's appearance augmentations and cropping are characteristic of state-of-the-art contrastive methods \cite{chen2020simple, he2020momentum}, \textit{making InsLoc a fitting contrastive case study}. 
            
            \subsubsection{Strategies to enhance InsLoc}
            
            We propose multiple strategies to augment the InsLoc view pipeline (\textit{s'}) for enhanced robustness in appearance-shifted and out-of-context detection scenarios. First, we consider cropping since InsLoc, like other contrastive methods \cite{chen2020simple, he2020momentum}, uses aggressive random cropping to create positive views (see Fig. \ref{intro_fig}). As random cropping has been shown to bias a model towards texture \cite{hermann2019origins}, we reason that InsLoc may struggle when texture shifts in detection such as when it is raining or snowing. Models that learn shape on the other hand can be effective in such situations \cite{geirhos2018imagenet}. We thus explore simple strategies to encourage InsLoc to learn shape. In particular, we experiment with \textit{geometric changes to crops}, specifically increasing the minimum \% of an image used to crop \textit{m} and enforcing an IoU constraint \textit{t} between views. We expect such changes to increase the spatial consistency between crops and encourage the model to learn object parts and shapes. In turn, the model can become more robust to texture shifts.
            
            \begin{figure}[t]
    \centering
    \includegraphics[scale=0.25]{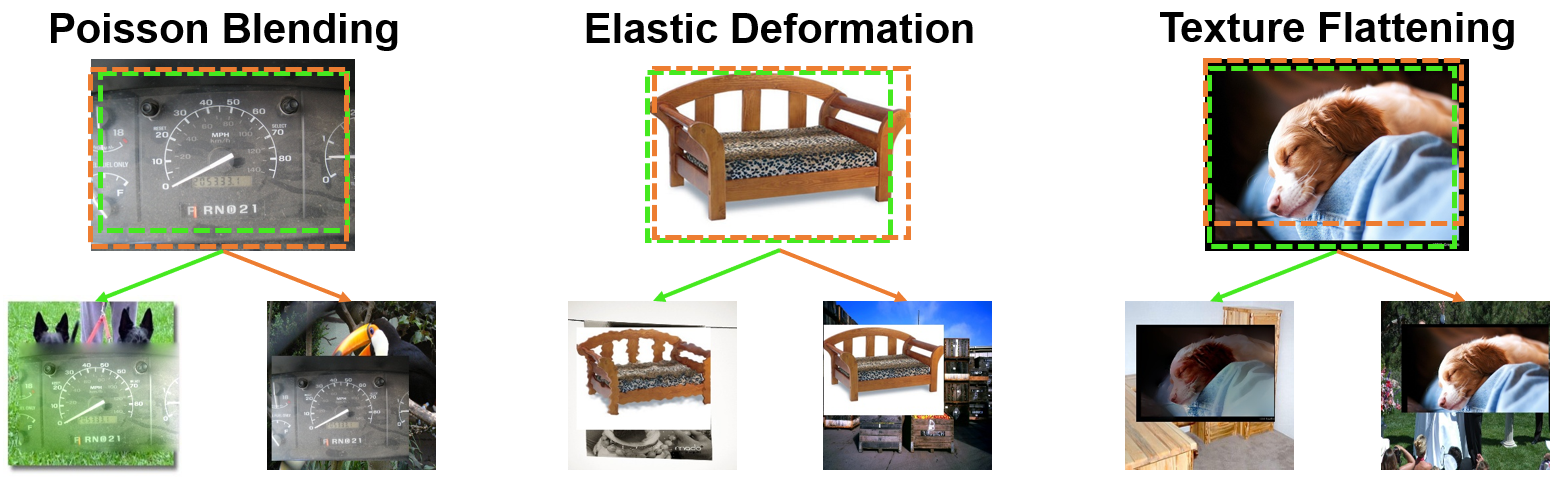}
    \caption{Our shortcut-reducing augmentations of study, shown within InsLoc. We perform each augmentation on the crop in the query view (left), but not in the key view (right).}
    \label{appear_aug}
\end{figure}

            Furthermore, we consider adding \textit{shortcut-reducing appearance augmentations}, as we find that InsLoc may not adequately discourage the model from attending to shortcuts that are non-robust in appearance-shifted scenarios, such as high-frequency noise and texture or color histograms. One strategy we explore is \textit{Poisson blending} \cite{perez2003poisson}, which is a method to seamlessly blend crops into a background image. We use Poisson blending instead of simple copy-pasting in InsLoc to reduce contrast between foreground and background regions, effectively making the pretext task harder as the model cannot use contrast as a shortcut to solve the task. It is also found that Poisson blending can introduce random illumination effects from the background, which may be desirable to learn invariance towards for appearance shifts. Second, we explore the \textit{texture flattening} application of solving the Poisson equation, as it washes out texture and changes brightness while only preserving gradients at edge locations. We reason that this augmentation can be effective to teach the model to not overfit to high-frequency texture shortcuts. Last, we investigate \textit{elastic deformation} \cite{simard2003best}, an augmentation that alters images by moving pixels with displacement fields. This augmentation can help make features more invariant to local changes in edges and noise shortcuts. We illustrate our proposed use of these strategies in Fig. \ref{appear_aug}. Augmentations are applied 100\% of the time, unless otherwise noted. We use Poisson blending and texture flattening as provided in the OpenCV library \cite{opencv_library} and the algorithm of \cite{simard2003best} for elastic deformation.  
            
            Lastly, we note that random cropping may result in the aligning of context (background and objects or objects and objects), which can lead to representations that are contextually biased and not robust in out-of-context detection. To address this problem, we experiment with \textit{strategies that use saliency-based object priors for crops}, as they can enable crops to refer to specific object regions rather than to background or co-occurring objects. In particular, we investigate two state-of-the-art approaches \cite{mo2021object, selvaraju2021casting}, as well as a hybrid of such approaches. We compare each strategy out-of-domain and also consider combining saliency strategies with shape and appearance strategies.

            \subsubsection{Pretraining and finetuning datasets}
            
             We identify two pretraining scenarios \textbf{s} to evaluate the robustness of contrastive view design strategies. First is ImageNet pretraining \cite{krizhevsky2012imagenet}, a standard scenario for contrastive approaches \cite{chen2020simple, he2020momentum}.  With the heavy computational nature of contrastive pretraining, along with our goal to conduct multiple experiments, we sample ImageNet from over 1 million to 50,000 images and call this set \textit{ImageNet-Subset}. Notably, most ImageNet images are \textit{iconic}, containing a single, large, centered object. For a dataset with different properties, we also consider pretraining on COCO \cite{lin2014microsoft}, which contains more \textit{scene} imagery, having multiple, potentially small objects. With the goal of self-supervised learning to learn robust representations on large, uncurated datasets, which are likely to contain scene imagery, COCO is a practical dataset to study. Also its multi-object nature makes it an apt test case for benchmarking contextual bias downstream. We pretrain specifically with COCO2017train and do not sample since its size is relatively small (118,287 images).
             
             We explore two finetuning datasets \textbf{d}: VOC  \cite{everingham2010pascal} when \textbf{s} is ImageNet-Subset and both VOC and COCO2017train when \textbf{s} is COCO. For VOC, we specifically use VOC0712train+val (16,551 images). We evaluate on COCO2017val (5,000 images) and VOC07test (4,952 images). In our sampling of ImageNet, we ensure semantic overlap with VOC by choosing 132 images for each of 379 classes from 13 synset classes that are related to VOC's classes: aircraft, vehicle, bird, boat, container, cat, furniture, ungulate, dog, person, plant, train, and electronics. 
            \begin{figure}[t]
    \centering
    \includegraphics[scale=0.24]{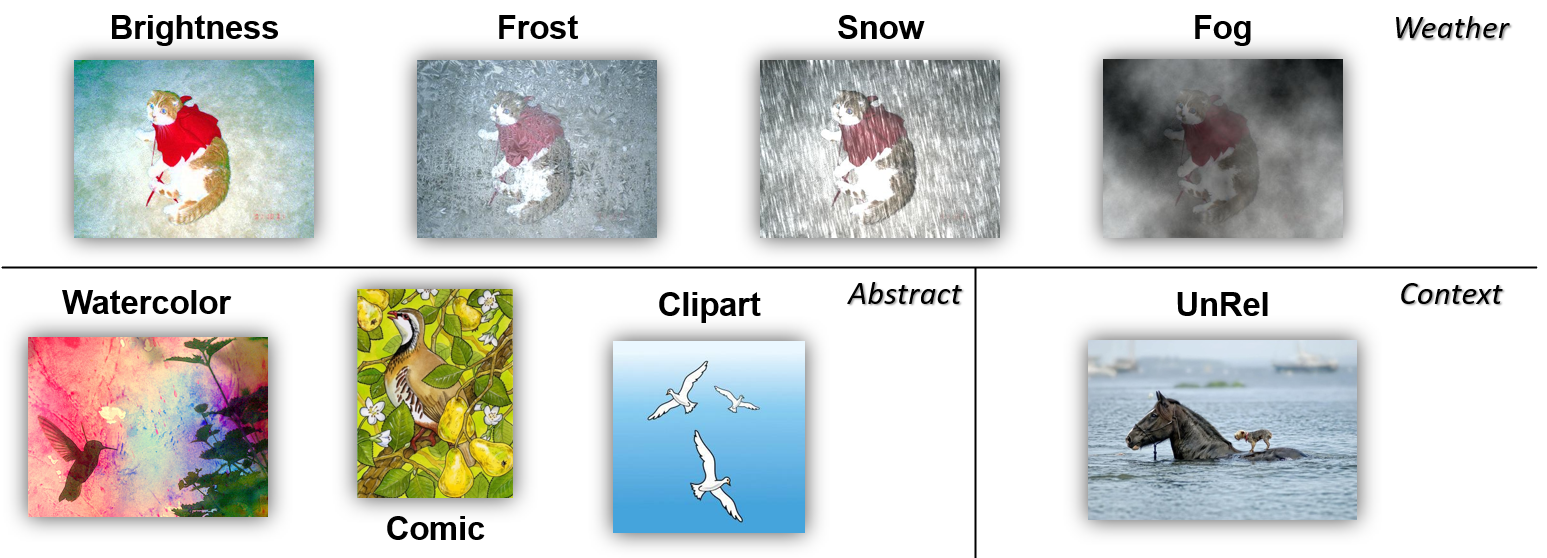}
    \caption{Our domain-shifted test sets. Weather and Abstract are for appearance shifts, and UnRel is for context shifts.}
    \label{ex_shifts}
\end{figure}

            \subsubsection{Domain shift datasets}

            We select various datasets $\mathbf{t_i}$ for out-of-domain evaluation. First, when \textbf{d} is VOC, we test on the challenging, abstract Clipart, Watercolor, and Comic object detection datasets \cite{inoue2018cross}, as they represent significant domain shifts in appearance. Clipart has the same 20 classes as VOC and 1,000 samples, while Watercolor and Comic share 6 classes with VOC and have 2,000 samples each. We take the average performance across the three sets and describe the overall set as \textit{Abstract}. When \textbf{d} is COCO, we test on the out-of-context UnRel dataset \cite{peyre2017weakly}. This set captures relations that are ``unusual"  between objects (\eg car under elephant), making this set useful for evaluating out-of-context robustness. We evaluate on 29 classes which overlap with COCO thing classes (1,049 images). Lastly, for both VOC and COCO, we consider Pascal-C and COCO-C \cite{michaelis2019benchmarking}, a collection of sets that are synthetically domain-shifted on natural corruption types. In particular, we explore the appearance-based Weather split at severity level 5, which consists of brightness, fog, frost, and snow shifts. We refer to the overall sets for VOC and COCO as VOC-Weather and COCO-Weather, respectively. Examples for the test sets are shown in Fig. \ref{ex_shifts}.

            \subsubsection{Training setup}
            
             Pretraining is performed with the provided InsLoc implementation \cite{yang2021instance}. Faster R-CNN \cite{ren2015faster}, with a ResNet-50 backbone and FPN, serves as the trained detector. With high computational costs for contrastive pretraining, these experiments consider a fixed pretraining budget of 200 epochs. For COCO, pretraining is performed with per-GPU batch size 64 and learning rate 0.03 on 4 NVIDIA Quadro RTX 5000 GPUs with memory 16 GB. For ImageNet-Subset, pretraining is performed with per-GPU batch size 32 and learning rate 0.015 on 2 NVIDIA GeForce GTX 1080 Ti GPUs with memory 11 GB. All pretraining uses a dictionary size of \textit{K}=8,192. Full finetuning of all layers is performed within the Detectron2 \cite{wu2019detectron2} framework with a 24k iteration schedule, a learning rate of 0.02, and a batch size of 4 on 2 NVIDIA GeForce GTX 1080 Ti GPUs, unless otherwise noted.

        \section{Experiments and Analysis}
        
            In this section, we outline various strategies for contrastive view design and evaluate their effectiveness in the InsLoc pretext task, considering both pretraining on ImageNet-Subset and COCO. Evaluation metrics are AP and AP$_{50}$. 
            
            \subsection{Pretraining on ImageNet-Subset}
            
            \subsubsection{How can we encourage contrastive learning to capture object shape and become more robust to appearance domain shifts?}
            
                \begin{figure}[t]
    \centering
    \includegraphics[scale=0.235]{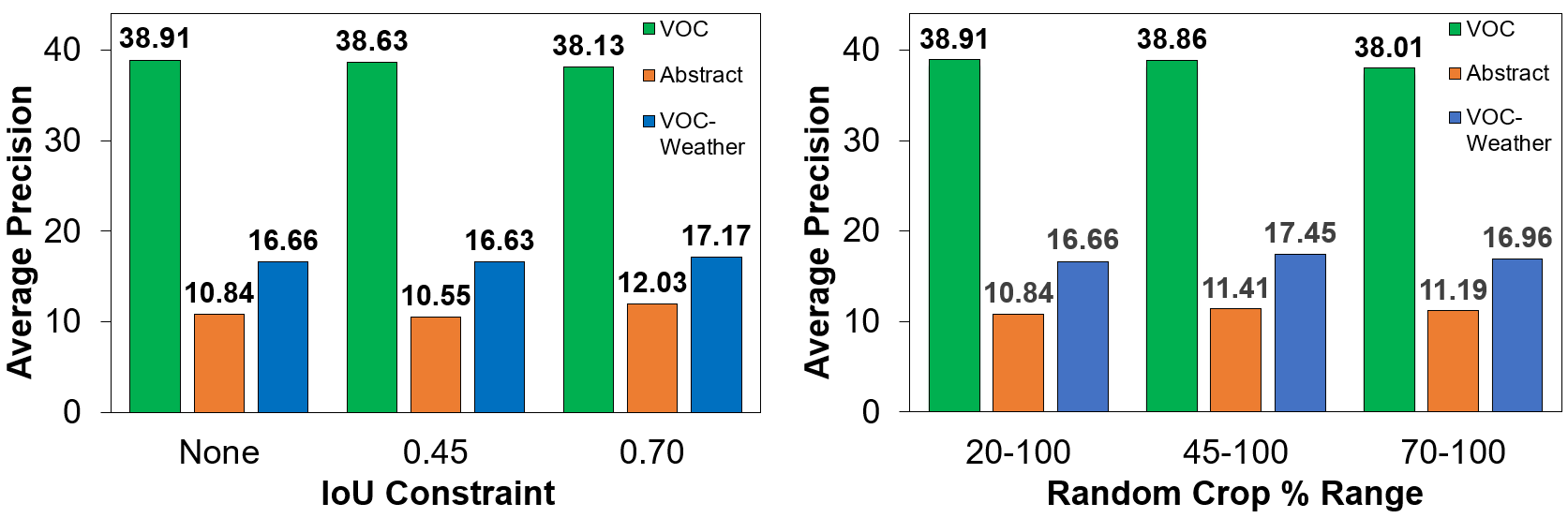}
    \caption{In-domain (VOC) vs. out-of-domain (Abstract, VOC-Weather) AP for models pretrained in InsLoc with various values of IoU constraint \textit{t} and minimum \% crop \textit{m}, averaged over three trials. Note that the top-performing settings for domain robustness (0.70 IoU constraint, 45-100\% crop) are different from the settings of the InsLoc baseline (20-100\% crop and no IoU constraint). }
    \label{crop_and_iou}
\end{figure}
                
                First, we consider appearance domain shifts in detection, where object shapes are preserved and texture is distorted. We wish to encourage InsLoc to capture object shapes for such scenarios and thus propose two simple strategies: (1) increasing the minimum \% of an image sampled as a crop \textit{m} and (2) adding an IoU contraint \textit{t}, such that query and key crops must have at least such IoU. To evaluate these strategies, we pretrain InsLoc on ImageNet-Subset using two different values of \textit{m}, 45\% and 70\%, in addition to InsLoc's default value of 20\%, while keeping the maximum crop bound as 100\%. InsLoc is additionally pretrained with two IoU constraint values of \textit{t}, 45\% and 70\%. Then for each experiment, we perform finetuning  on VOC and evaluate in-domain on VOC and out-of-domain on Abstract and VOC-Weather (two sets with distorted texture). 
                
                Results over three trials are shown in Fig. \ref{crop_and_iou}. Notably, we find that the default InsLoc settings (20-100\% crops, no IoU constraint) result in the best in-domain AP, but \textit{not the top} out-of-domain AP. In particular, \textit{m}$=$45 and \textit{t}$=$70 have the highest out-of-domain AP for their respective value comparisons (up to +1.19 AP on Abstract, +0.79 AP on VOC-Weather). In general, these results show out-of-domain benefits with having substantial overlap and higher minimum crop \%. \textit{Overall, these results highlight that including larger object regions in crops and encouraging spatial consistency between views are effective strategies to ensure greater robustness to appearance shifts.} We also note that the robustness sweet spot for \textit{m} may be related to an observed tradeoff with in-domain AP, which drops as \textit{m} or \textit{t} increases. We reason that while we are encouraging the model to learn shape features, we are also increasing the probability that the model can attend to natural, high-frequency shortcuts in images since crops that have more area and overlap more might share more of these signals. We next consider augmentations to remove shortcuts and improve these strategies.


            \subsubsection{Can shortcut-reducing augmentations make shape strategies more effective?}
            
                Though contrastive view pipelines typically have significant appearance augmentations like Gaussian blur, grayscale conversion, and color jitter \cite{chen2020improved}, we reason that even more aggressive augmentations may be beneficial with our \% crop and IoU strategies to further limit shortcuts and better learn shape. SimCLR \cite{chen2020simple} serves as motivation, as the authors explore augmentations to avoid color histogram shortcuts. We explore Poisson blending, texture flattening, and elastic deformation as stronger augmentations to similarly reduce shortcuts and enable InsLoc to learn robust features.
                
                \begin{table}[t]
    \begin{center}
        \renewcommand{\arraystretch}{1}
        \begin{tabular}{|c|g|c|c|}
            \hline
                \textbf{\small Method}  & \multicolumn{1}{c|}{\textbf{\small VOC}} & \multicolumn{1}{c|}{\textbf{\small Abstract}} & \multicolumn{1}{c|}{\textbf{\small Weather}} \\
                 & \multicolumn{1}{c|}{\small AP} &  \multicolumn{1}{c|}{\small AP} &  \multicolumn{1}{c|}{\small AP} \\
                \hline\hline
                \small InsLoc, \textit{m}=20 (Baseline) & \small
                38.91 & \small 10.84 & \small 16.66 \\
                \hline
       \small +Poisson Blending, \textit{m}=20  & \small 38.49 & \small 11.95 & \small 16.98  \\
                \hline\hline
                    \small InsLoc, \textit{m}=45  & \small 38.86 & \small 11.41 & \small 17.45 \\
                \hline
                     \small +Poisson Blending, \textit{m}=45  & \small 40.22 & \small 12.96 & \small 19.23  \\
                \hline\hline
                
                \small InsLoc, \textit{m}=70  & \small 38.01 & \small 11.19 & \small 16.96 \\
                \hline
                
                     \small +Poisson Blending, \textit{m}=70  & \small \textbf{40.54} & \small \textbf{13.00} & \small \textbf{19.73} \\ \hline\hline
                \small InsLoc, \textit{t}=70  & \small 38.13 & \small 12.03 & \small 17.17 \\
                
                \hline
                 \small \small +Poisson Blending, \textit{t}=70 & \small 39.31 & \small 11.97 & \small 18.36 \\
                \hline
        \end{tabular}
    \end{center}
    \caption{In-domain (VOC) vs. out-of-domain (Abstract, VOC-Weather) AP following InsLoc pretraining with and without Poisson blending InsLoc's query crop, for various values of min \% crop \textit{m} and IoU threshold \textit{t}. Note that the use of our Poisson blending strategy results in substantial in-domain and out-of-domain gains, especially at \textit{m}=70.}
    \label{pois}
\end{table}

                \begin{table}[t]
    \begin{center}
        \renewcommand{\arraystretch}{1}
        \begin{tabular}{|c|g|c|c|}
            \hline
                \textbf{\small Method}  & \multicolumn{1}{c|}{\textbf{\small VOC}} & \multicolumn{1}{c|}{\textbf{\small Abstract}} & \multicolumn{1}{c|}{\textbf{\small Weather}} \\
                 & \multicolumn{1}{c|}{\small AP} &  \multicolumn{1}{c|}{\small AP} &  \multicolumn{1}{c|}{\small AP} \\
                \hline\hline
                \small InsLoc, \textit{m}=20 (Baseline)  & \small 38.91 & \small 10.84 & \small 16.66 \\
                \hline
            
                \small InsLoc, \textit{m}=70  & \small 38.01 & \small 11.19 & \small 16.96 \\
        
                \hline
                \hline
                 \small +Poisson Blending, \textit{m}=70  & \small 40.54 & \small 13.00 & \small \textbf{19.73}  \\

                \hline
                 \small +Elastic Deformation, \textit{m}=70  & \small 40.94 & \small {13.26} & \small 18.70 \\
                \hline
                 \small +Texture Flattening, \textit{m}=70  & \small 40.45 & \small \textbf{13.57} & \small 19.58  \\
                \hline
                \hline
                 \small \small  +Apply 25\% of Time, \textit{m}=70 & \small \textbf{41.64} & \small 12.53 & \small 19.70  \\
                \hline
        \end{tabular}
    \end{center}
    \caption{In-domain (VOC) vs. out-of-domain (Abstract, VOC-Weather) AP when pretraining InsLoc with shortcut-reducing augmentations at min \% crop \textit{m}=70. ``Apply 25\% of Time" means that we either apply Poisson blending, apply elastic deformation, apply texture flattening, or use the baseline setting, each with probability 25\%.}
    \label{appearaugs_table}
\end{table}
                
                 To first test how augmentations interact with the \% crop and IoU strategies, we perform Poisson blending with various values of \textit{m} and \textit{t}, shown in Table \ref{pois}. Different from in Fig. \ref{crop_and_iou}, we find that the top domain robustness setting is \textit{m}=70 (rather than \textit{t}=70) and that \textit{significant} out-of-domain gains over the InsLoc baseline are achieved in such setting (+2.16 AP on Abstract, +3.07 AP on VOC-Weather). Moreover, we find that \textit{in-domain} AP also increases in this setting (+1.63 AP), indicating that \textit{shortcut augmentations can enable the learning of shape without tradeoffs in-domain}. Note also that Poisson blending at \textit{m}=20 is not effective in-domain and is less effective out-of-domain. \textit{These results indicate that shortcut-reducing augmentations may not be effective unless the model is encouraged to capture robust object features like shape}, which can be done at higher crop \%.
                
                We further use the top setting of \textit{m}=70 to test each of Poisson blending, elastic deformation, and texture flattening, shown over three trials in Table \ref{appearaugs_table}. We find that all augmentations help both in-domain and out-of-domain. In particular, Poisson blending is the top for VOC-Weather (+3.07 AP) and texture flattening is for Abstract (+2.73 AP). We reason that texture flattening simulates the flattened texture of digital Abstract imagery well, while Poisson blending's random illumination effects are helpful for the texture changes seen with weather. Also shown in Table \ref{appearaugs_table} is a scenario where we either apply one of the three augmentations or just use the default InsLoc setting, each with probability 25\%. We find even more substantial gains in-domain (+2.73 AP) in such scenario. \textit{These results demonstrate that the benefits of creating views  with shape-encouraging and shortcut-reducing strategies are not limited to out-of-domain robustness, and these strategies can lead to more robust object features overall.} To our knowledge, we are the first to demonstrate such effectiveness of Poisson blending, texture flattening, and elastic deformation as augmentations for contrastive views.

            \subsection{Pretraining on COCO}

             \subsubsection{How do saliency-based view strategies compare on out-of-context and appearance domain shifts?}
                      
                Intra-image cropping in contrastive learning has been noted to be potentially  harmful when pretraining on multi-object images (\eg  COCO) \cite{purushwalkam2020demystifying}. Approaches have aimed to reduce the impact of contextual bias in such case through using saliency-based object priors, with OA-Crop \cite{mo2021object} and CAST \cite{selvaraju2021casting} representing two more robust cropping methods. OA-Crop uses an initial pretraining of MoCo-v2 to gather Contrastive Class Activation Maps and creates a number of object crops for an image from bounding boxes around salient regions from these maps. During training, one randomly selected object crop, rather than the entire image, is used as the source from which to crop views. CAST alternatively ensures that crops overlap with saliency maps gathered with DeepUSPS \cite{nguyen2019deepusps}, an ``unsupervised" saliency detector (though it still uses ImageNet-supervised weights). 
                \begin{figure}[t]
    \centering
    \includegraphics[scale=0.31]{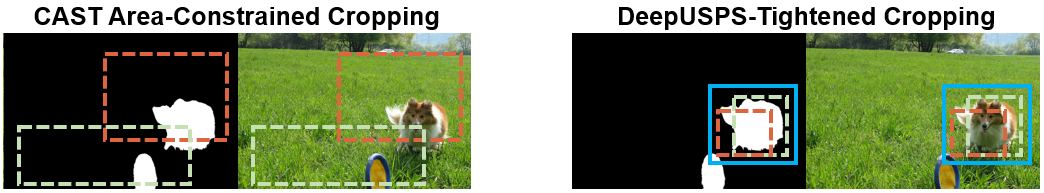}
    \caption{CAST \cite{selvaraju2021casting} vs. DeepUSPS-Tightened crops. The blue box shows a randomly chosen object crop that is the source for query and key views. Green and orange boxes show example query and key crops.}
    \label{cast_ex}
\end{figure}
                        
                        \begin{table}[t]
    \begin{center}
        \renewcommand{\arraystretch}{1}
        \begin{tabular}{|c|g|c|c|}
            \hline
                \textbf{\small Cropping Method}  & \multicolumn{1}{c|}{\textbf{\small COCO}} & \multicolumn{1}{c|}{\textbf{\small UnRel}} & \multicolumn{1}{c|}{\textbf{\small Weather}} \\
                 & \multicolumn{1}{c|}{\small AP$_{50}$} &  \multicolumn{1}{c|}{\small AP$_{50}$} &  \multicolumn{1}{c|}{\small AP$_{50}$} \\
                \hline\hline
                \small InsLoc & \small 26.16 & \small 22.60 & \small 10.53 \\

                \hline
                \hline
                 \small +OACrop & \small 25.55 & \small 23.10 & \small 9.82  \\
                \hline
                 \small +CAST & \small 26.70 & \small {22.36} & \small 10.58 \\
                 \hline
                \small +DeepUSPS-Tight, \textit{m}=8 & \small 27.76 & \small 24.59 & \small 12.20  \\
                \hline
                 \small +DeepUSPS-Tight, \textit{m}=20 & \small \textbf{28.39} & \small \textbf{26.11} & \small \textbf{12.33}  \\
                \hline
        \end{tabular}
    \end{center}
    \caption{In-domain (COCO) vs. out-of-domain (UnRel, COCO-Weather) AP$_{50}$ of saliency strategies within InsLoc. For InsLoc, OACrop, and CAST, results are with the default or optimal cropping values if reported (\textit{m}=20, 8, and 20 respectively). DeepUSPS-Tightened is tested at \textit{m}=8 and 20.}
    \label{context_compare}
\end{table}
                Notably, these methods have not been evaluated in out-of-domain detection, so we fill in this gap by comparing them within InsLoc. We also consider a hybrid approach called \textit{DeepUSPS-Tightened} crops, where DeepUSPS saliency maps, rather than ContraCAMs, are used to create object crops like OA-Crop, as we observe DeepUSPS's maps are higher quality. We emphasize that the difference between between CAST and DeepUSPS-Tightened crops is that maps are used with CAST to ensure that crops \textit{overlap} with objects, rather than to \textit{reduce} background area and \textit{tighten} crops to objects, which is the goal of the hybrid that we propose. We exemplify these differences in Fig. \ref{cast_ex}.
                
                In Table \ref{context_compare}, we compare each strategy following COCO pretraining and finetuning in terms of AP$_{50}$ on COCO, the out-of-context UnRel, and the appearance-shifted COCO-Weather. We use the saliency maps provided by \cite{mo2021object} and \cite{selvaraju2021casting}, as well as their top reported min \% crop (or default if not reported). We also test DeepUSPS-Tightened at  \textit{m}=8 and \textit{m}=20. A first observation is that there is a \textit{significant} domain gap (-3.56 AP$_{50}$) between COCO and UnRel without incorporating any saliency strategy, \textit{indicating contextual bias in downstream object detection}. The OACrop strategy improves AP$_{50}$ on UnRel, while CAST does not. Alternatively, CAST improves on COCO and COCO-Weather (slightly) while OA-Crop does not. Notably, \emph{our hybrid DeepUSPS-Tightened leads to the top gains} vs. InsLoc (+2.23 AP$_{50}$ on COCO, +3.51 AP$_{50}$ on UnRel, +1.80 AP$_{50}$ on COCO-Weather). The domain gap is much smaller than InsLoc as well (-2.28 vs. -3.56 AP$_{50}$).
                
          \begin{figure}[t]
    \centering
    \includegraphics[scale=0.26]{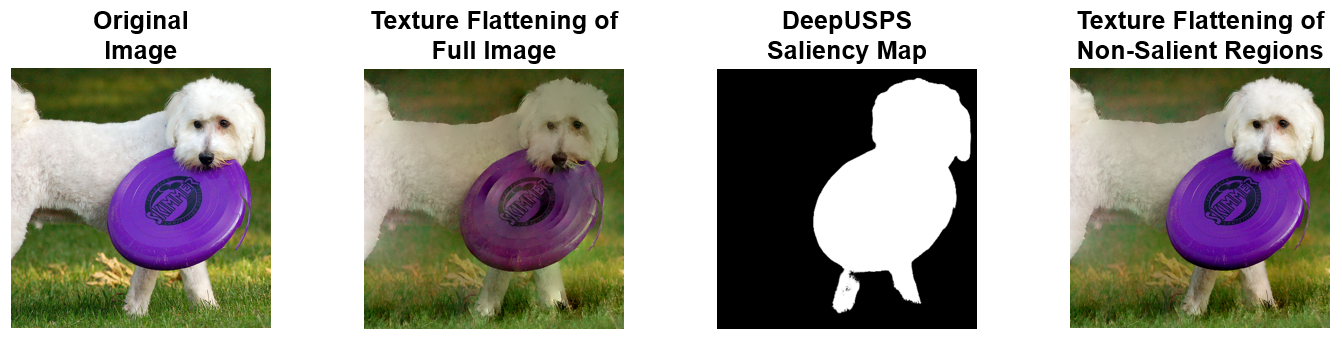}
    \caption{Our proposed texture flattening of non-salient regions (TFNS) augmentation vs. full image texture flattening.}
    \label{tfns}
\end{figure}
                 \begin{table}[t]
            \begin{center}
                \renewcommand{\arraystretch}{1}
                \begin{tabular}{|c|g|c|c|}
                    \hline
                        \small \textbf{Method} & \multicolumn{1}{c|}{\small \textbf{COCO}} & \small \textbf{UnRel} & \small \textbf{Weather} \\
                         & \multicolumn{1}{c|}{AP$_{50}$} & AP$_{50}$ & AP$_{50}$   \\
                        \hline\hline
                        \small InsLoc  & \small 26.16 & \small 22.60 &  \small 10.53 \\
                        \hline
                        \hline
                        \small+DeepUSPS-Tight/TF & \small 27.73 & \small 25.27 & \small 11.94 \\
                        \hline
                         \small+DeepUSPS-Tight/TFNS  & \small\textbf{28.74} & \small\textbf{25.71} & \small \textbf{12.44} \\
                        \hline
                \end{tabular}
            \end{center}
            \caption{In-domain (COCO) vs. out-of-domain (UnRel, COCO-Weather) AP$_{50}$ for full image (TF) vs. non-salient region texture flattening (TFNS) of InsLoc query crops in combination with DeepUSPS-Tightened strategy (at \textit{m}=70), with respect to InsLoc baseline (at \textit{m}=20). }
            \label{tf_vs_tfns}

        \end{table}

        \begin{table*}[t]
    \begin{center}
        \renewcommand{\arraystretch}{1}
        \begin{tabular}{|c|c|c||g|c|c||g|c|c|}
            \hline
                 &  &
                 & \multicolumn{3}{c||}{\small \textit{VOC Finetuning}} & \multicolumn{3}{c|}{\small \textit{COCO Finetuning}} \\
                \textbf{\small Method} & \textbf{\small Crop \%} &
                \textbf{\small Augmentations} &  \multicolumn{1}{c}{\small \textbf{VOC}} & \multicolumn{1}{c}{\small \textbf{Abstract}} & \multicolumn{1}{c||}{\small \textbf{Weather}} &  \multicolumn{1}{c}{\small \textbf{COCO}} & \multicolumn{1}{c}{\small \textbf{UnRel}} & \multicolumn{1}{c|}{\small \textbf{Weather}}\\
                & & & \multicolumn{1}{c}{\small AP} & \multicolumn{1}{c}{\small AP} & \multicolumn{1}{c||}{\small AP} & \multicolumn{1}{c}{\small AP$_{50}$} & \multicolumn{1}{c}{\small AP$_{50}$} & \multicolumn{1}{c|}{\small AP$_{50}$}  \\
                \hline\hline
                \small InsLoc  
                & \small 20-100 & \small Default & \small 39.79 & \small 11.31 & \small 18.47 & \small 26.16 & \small 22.60 & \small 10.53   \\
                & \small 70-100 & \small Default  &  \small 38.45 & \small 12.19  & \small 17.37 & \small 24.70 & \small 20.87 & \small 9.56 \\
                & \small 20-100 & \small +TFNS & \small \underline{40.43}  & \small 12.45 & \small \underline{19.42} & \small 26.94 & \small 23.00 & \small \underline{11.37}  \\
                & \small 70-100 & \small +TFNS & \small 40.09 & \small \underline{12.56}  & \small 19.05 &  \small \underline{27.12} & \small \underline{23.03} & \small 11.25\\

                \hline
                \small InsLoc  & \small 8-100 & \small Default & \small 40.87  & \small 11.03 & \small 19.85 & \small 27.76 & \small 24.59 & \small 12.20   \\
               \small  +DeepUSPS-Tightened& \small 20-100 & \small Default & \small 41.56  & \small 11.78 &\small 19.86  & \small28.39 & \small26.11 &\small 12.33 \\
               & \small 70-100 & \small Default &\small 39.30 &\small 12.11 &\small 17.69 &\small 25.46 &\small 22.66 & \small10.67 \\
                & \small8-100 & \small +TFNS &\small 41.08 & \small12.48 & \small \textbf{\underline{20.64}} & \small 28.44 & \small \textbf{\underline{27.31}} & \small 12.27  \\
                & \small 20-100 & \small +TFNS & \small 41.70  &\small 12.50 & \small 19.54 & \small 28.66 & \small 24.93 & \small 12.12  \\
                & \small 70-100 & \small +TFNS & \textbf{\small \underline{41.80}} & \textbf{\small \underline{13.18}} &\small 20.46 & \textbf{\small \underline{28.74}} & \small25.71 & \textbf{\small \underline{12.44}}   \\
                \hline
        \end{tabular}
    \end{center}
    \caption{InsLoc+DeepUSPS-Tightened vs. InsLoc pretrained at various crop \% and with/without texture flattening of non-salient regions (TFNS), using 24k schedule for both VOC and COCO finetuning. Underlined=top per method, bold=best overall. }
    \label{main_voc_24k}
\end{table*}

                We infer that the high quality of DeepUSPS maps along with the \textit{removal} of background through cropping are potential reasons for  DeepUSPS-Tightened to have the top AP$_{50}$. In detection pretraining, we reason that it may not be important to include background in crops, and that \textit{strategic use of a quality saliency detector can enable detectors to be less biased and more robust out-of-context.} 
             
             \subsubsection{Can shape and shortcut-reducing strategies further help robustness with saliency strategies?}
             
             While tightened crops remove many non-salient pixels, some remain inside crops. The pretext task can thus be solved by matching positives based on the background, rather than objects, which can hurt out-of-context robustness. Moreover, aggressive cropping of even object-based crops can still lead to representations that do not capture shape well, hurting appearance shift robustness. Inspired by the ImageNet-Subset experiments, we also explore shape and appearance strategies with COCO pretraining. In particular, we consider \textit{m}=70 as a shape strategy and texture flattening as a shortcut-reducing strategy. Since we have saliency maps, we also propose another strategy: texture flattening of non-salient regions only (TFNS). We specifically propose to distort the background (non-salient regions marked by DeepUSPS) of one InsLoc view to encourage background invariance between views during pretraining. We reason that TFNS may be effective for out-of-context robustness as shortcuts may come more significantly from the background rather than salient object regions. We illustrate this proposed augmentation in Fig. \ref{tfns}.

            In Table \ref{tf_vs_tfns}, we show results with \textit{m}=70, DeepUSPS-Tightened crops, and texture flattening (full and just non-salient regions). We observe that both strategies are effective, and TFNS leads to larger gains across all sets. We reason that learning some level of texture, even with shape, is important, and TFNS preserves important texture (of objects) while removing unimportant texture (high-frequency shortcuts which come from the background).

            For a more thorough evaluation of TFNS and DeepUSPS-Tightened, in Table \ref{main_voc_24k}, we present results of InsLoc with these strategies and various values of \textit{m}, in finetuning on both VOC and COCO. We find that the combination of DeepUSPS-Tightened, \textit{m}=70, and TFNS results in the top AP on VOC and Abstract and the top AP$_{50}$ on COCO and COCO-Weather. \textit{These results illustrate the high overall effectiveness of combining shape, shortcut-reducing, and saliency strategies.} We also observe that the top performance on UnRel (27.31 AP$_{50}$) is achieved at \textit{m}=8, along with TFNS and DeepUSPS-Tightened. We reason that since context is a ``natural" domain shift, where object texture is preserved, shape may less useful, and aggressive cropping at \textit{m}=8 of \textit{object crops with mostly salient pixels} can result in effective texture features. High performance is even achieved on VOC-Weather in this setting, demonstrating these features to be robust to some texture shift. A last note is that we find TFNS gains to be highest at \textit{m}=70, which makes sense as many non-salient pixels exist in such views.

            \subsubsection{How does texture flattening of non-salient regions compare to another strategy for context debiasing?}
            
                We gain further understanding of the effectiveness of TFNS through evaluating it versus replacing the query crop's non-salient pixels with a random grayscale value, a top background debiasing strategy \cite{ryali2021learning,zhao2021distilling}. Results are shown for COCO pretraining and VOC finetuning in Table \ref{tf}. We find that TFNS outperforms the grayscale strategy on all sets. We reason that TFNS is more beneficial for background debiasing as in distorting the background, it maintains continuity between an image's salient and non-salient pixels, making images seen in pretraining more natural and closer to those seen at test time.
                   \begin{table}
    \begin{center}
        \renewcommand{\arraystretch}{1}
        \begin{tabular}{|c|g|c|c|}
            \hline
                \textbf{\small Method} & \multicolumn{1}{c|}{\small \textbf{VOC}} & \multicolumn{1}{c|}{\small \textbf{Abstract}} & \multicolumn{1}{c|}{\small \textbf{Weather}} \\
                & \multicolumn{1}{c|}{\small AP} & \small AP & \small AP \\
                \hline\hline
               \small InsLoc+RandGrayBG & \small 40.99 &\small 12.97 &\small 18.36 \\
                \hline
                \small InsLoc+TFNS  & \small \textbf{41.80} & \small \textbf{13.18} & \small \textbf{20.46} \\
                \hline
        \end{tabular}
    \end{center}
    \caption{InsLoc+DeepUSPS-Tightened (\textit{m}=70), using query crops with random grayscale backgrounds (RandGrayBG) vs. crops with non-salient region texture flattening (TFNS).}
    \label{tf}
\end{table}
                \begin{table}[t]
    \begin{center}
        \renewcommand{\arraystretch}{1}
        \begin{tabular}{|c|gg|cc|cc|}
            \hline
                \textbf{\small Method}  & \multicolumn{2}{c|}{\textbf{\small COCO}} & \multicolumn{2}{c|}{\textbf{\small UnRel}} & \multicolumn{2}{c|}{\textbf{\small Weather}} \\
                 & \multicolumn{1}{c}{\small AP} & \multicolumn{1}{c|}{\small AP$_{50}$} & \multicolumn{1}{c}{\small AP} & \multicolumn{1}{c|}{\small AP$_{50}$} & \multicolumn{1}{c}{\small AP} & \multicolumn{1}{c|}{\small AP$_{50}$} \\
                \hline\hline
                \small InsLoc & \small 29.63 & \small 46.50 & \small 25.88 & \small 41.68 & \small 14.25 & \small 23.32 \\
                \hline
                 \small Ours  & \small \textbf{30.08} & \small \textbf{47.30} & \small \textbf{27.46} & \small \textbf{42.93} & \small \textbf{14.52} & \small \textbf{23.81} \\
                \hline
        \end{tabular}
    \end{center}
    \caption{InsLoc with our top strategies (DeepUSPS-Tightened, \textit{m}=70, and texture flattening of non-salient regions) vs. InsLoc baseline. Results are compared following COCO pretraining and finetuning (using a 2$\times$ schedule).}
    \label{2x}
\end{table}
             
            \subsubsection{How are results at a longer training schedule?}
            
                In Table \ref{2x}, we lastly show that our strategies maintain effectiveness on COCO even when using a longer finetuning schedule (the 2$\times$, 180k iteration schedule from \cite{yang2021instance}). Gains can notably be observed on \textit{both} out-of-domain test sets: +1.58 AP on UnRel and +0.27 AP on COCO-Weather.

    \section{Conclusion}
    
        In this work, we present contrastive view design strategies to improve robustness to domain shifts in object detection. We show that we can make the contrastive augmentation pipeline more robust to domain shifts in appearance through encouraging the learning of shape (with higher minimum crop \% and IoU constraints). Furthermore, combining these shape strategies with shortcut-reducing appearance augmentations is shown to lead to more robust object features overall, demonstrated by both in-domain and out-of-domain performance improvements. Finally, when pretraining on multi-object image datasets with saliency map priors, we find that tightening crops to salient regions, along with texture flattening the remaining non-salient pixels in a view, is an effective strategy to achieve out-of-context detection robustness. Overall, these strategies can serve to guide view design in future detection-focused, contrastive pretraining methods.
        
        \noindent \textbf{Acknowledgements:} This work was supported by the National Science Foundation under Grant No. 2006885.


\bibliography{aaai23}

\bigskip

\end{document}